\documentclass[a4paper,twoside]{article}

\usepackage{graphicx}
\usepackage{subfigure}
\usepackage{calc}
\usepackage{amssymb}
\usepackage{amstext}
\usepackage{amsmath}
\usepackage{amsthm}
\usepackage{multicol}
\usepackage{pslatex}
\usepackage{SCITEPRESS}
\usepackage[small]{caption}
\usepackage{tikz}
\usepackage{pgfplots}
\usepackage{units}
\usepackage{natbib}
\usepackage{filecontents}

\usepackage{color}

\begin{document}
\setlength\parindent{0pt}

\title{Efficient Implementation of a Recognition System Using the Cortex Ventral Stream Model}

\author{\authorname{Ahmad Bitar, Mohammad M. Mansour and Ali Chehab}
\affiliation{Department of Electrical and Computer Engineering, American University of Beirut, Beirut 1107 2020, Lebanon}
\email{\{ab76, mmansour, chehab\}@aub.edu.lb}
}
\keywords{HMAX, Support Vector Machine, Nearest Neighbor, Caltech101}

\abstract{\small In this paper, an efficient implementation for a recognition system based on the original HMAX model of the visual cortex is proposed. Various optimizations targeted to increase accuracy at the so-called layers S1, C1, and S2 of the HMAX model are proposed.
At layer S1, all unimportant information such as illumination and expression variations are eliminated from the images. Each image is then convolved with 64 separable Gabor filters in the spatial domain. At layer C1, the minimum scales values are exploited to be embedded into the maximum ones using the additive embedding space. At layer S2, the prototypes are generated in a more efficient way using Partitioning Around Medoid (PAM) clustering algorithm.
The impact of these optimizations in terms of accuracy and computational complexity was evaluated on the Caltech101 database, and compared with the baseline performance using support vector machine (SVM) and nearest neighbor (NN) classifiers. The results show that our model provides significant improvement in accuracy at the S1 layer by more than 10\% where the computational complexity is also reduced. The accuracy is slightly increased for both approximations at the C1 and S2 layers.}

\abstract{\small In this paper, an efficient implementation for a recognition system based on the original HMAX model of the visual cortex is proposed. Various optimizations targeted to increase accuracy at the so-called layers S1, C1, and S2 of the HMAX model are proposed.
At layer S1, all unimportant information such as illumination and expression variations are eliminated from the images. Each image is then convolved with 64 separable Gabor filters in the spatial domain. At layer C1, the minimum scales values are exploited to be embedded into the maximum ones using the additive embedding space. At layer S2, the prototypes are generated in a more efficient way using Partitioning Around Medoid (PAM) clustering algorithm.
The impact of these optimizations in terms of accuracy and computational complexity was evaluated on the Caltech101 database, and compared with the baseline performance using support vector machine (SVM) and nearest neighbor (NN) classifiers. The results show that our model provides significant improvement in accuracy at the S1 layer by more than 10\% where the computational complexity is also reduced. The accuracy is slightly increased for both approximations at the C1 and S2 layers.}

\onecolumn \maketitle \normalsize \vfill

\section{\uppercase{Introduction}}\label{sec:introduction}

\noindent The human visual system is very powerful; it can recognize and differentiate among numerous similar objects in a very selective, robust and fast manner. Modern computers are able to translate the human ventral visual pathway (known as the ``WHAT" stream) in order to achieve, in a similar manner to the human brain, an impressive trade-off between selectivity and invariance. Several scientists have attempted to model and mimic the human vision system \citep{x6}.

The Hierarchical Model And X (HMAX) is an important model for object recognition in the visual cortex known for its high ability to achieve performance levels close to the human object recognition capability \citep{x1}. HMAX divides the human ventral stream into five layers: S1, C1, S2, C2 and View-TUned (VTU).

The first layer S1 of the HMAX model relies on the Gabor filter \citep{x2}, which  is a linear filter used for edge detection.~It differs from other filters by its capability to highlight all the features that are oriented in the direction of the filtering. The features are therefore extracted from the images by tuning the gabor filter to several different scales and orientations using fine-to-coarse approach.

Several methods have been proposed in the literature in order to improve the efficiency of the original HMAX model. \citet{x3} proposed an extension of the original HMAX model, emphasizing the importance of shape selectivity in area V4. A simpler radial basis function (RBF) model for object recognition was proposed by \citet{x4} to maintain a good degree of translation and scale invariance. The proposed model was considered better than the original HMAX for translation and scale invariance by changing the point of attention and decreasing the amount of visual information to be processed.  \citet{x5} developed a new set of receptive field shapes and parameters for cells in the S1 and C1 layers. The method serves to increase position invariance in contrast to scale invariance, which is decreased. \citet{x7} proposed a general framework for robust object recognition of complex visual scenes based on a quantitative theory of the ventral pathway of visual cortex. A number of improvements to the base model were proposed by \citet{x8} in order to increase the sparsity. The proposed model has shown a remarkable improvement on classification performance and the resulting model is found more economical in terms of computations.  \citet{x9} proposed several approximations at the four HMAX layers (S1, C1, S2 and C2) in order to increase the efficiency of the model in terms of accuracy and computational complexity. \citet{x10} proposed a semi-supervised learning algorithm for visual object categorization by exploiting unlabelled data and employing a hybrid generative-discriminative learning scheme. The method achieved good performance in multi-class object discrimination tasks.  \citet{x11} proposed a scheme based on a kernel function for discriminative classification. The method achieved improved accuracy and reduced computational complexity compared to the baseline model.

In this paper, the goal is to perform various optimizations at the S1, C1 and S2 layers of the original HMAX model. The results demonstrate that these optimizations increase the accuracy of the HMAX model as well as reduce its computational complexity at the S1 layer. The accuracy of the final model proves the advantage of exploiting only the important features for recognition and generating the prototypes in a more efficient way.

The remainder of this paper is organized as follows. In section 2,  a brief overview of the original HMAX model is explained. The proposed approximations at S1, C1 and S2 layers are presented in section 3, 4 and 5, respectively. Experimental results are shown in section 6. Finally, section 7 gives concluding remarks and some directions for future work.

\section{\uppercase{HMAX Model with Feature Learning}}

\noindent HMAX \citep{x12} is a computational model that summarizes the organization of the first few stages of object recognition in the WHAT pathway of the visual cortex, which is located in the occipital lobe at the back of the human brain. It is considered a primordial part of the cerebral cortex responsible for processing visual information in the first \unit[100-150]{ms}. Indeed, light enters our eye from the central aperture, called ``Pupil", and then passes through the ``Crystalline lens" which is considered the biconvex transparent body situated behind the iris into the eye and aiming to focus light on the retina that sends images to a specific part of the brain (visual cortex) through the optic nerve. The retina contains five different types of connected neurons: Photoreceptors (95\% rods and 5\% cones), Horizontal, Bipolar, Amacrine and Ganglion through which the light leaves the eye. The visual cortex, located in and around the calcarine sulcus, refers to the striate cortex V1, anatomically equivalent to Brodmann area 17 ($B A17$), connected to several extrastriate visual cortical areas, anatomically equivalent to Brodmann area 18 and Brodmann area 19. The right and left V1 receive information from the right and left Lateral Geniculate Nucleus (LGN), respectively. The LGNs are located in the thalamus of the brain and they receive information directly from the ganglion cells of the retina via the optic nerve and optic chiasm.

\subsection{Computational Complexity}
The operations of the five layers of the HMAX model are briefly summarized.

{\bf S1 layer:} All the responses of the S1 units are summarized here by simply performing 2-D convolution between 64 Gabor filters (16 scales in steps of two pixels  and 4 orientations) shown in Figure 1 and the input images in the spatial domain.\\
Firstly, each Gabor filter of a specific scale and orientation can be initialized as:
\begin{equation}\label{eq:gabor_equation_G}
\mathrm{G}(x,y)=\exp^{-\left(\frac{u^2 + \gamma^2v^2} { 2 \sigma^2}\right)} \times \cos \left(\frac{2 \pi} {\lambda} u\right),
\end{equation}
where:
\begin{align*}
    u &= xcos \theta + y sin \theta, \\
    v &= -x sin\theta + y cos\theta,\\
    \gamma & = 0.0036 \times \rho^2 + 0.35 \times \rho + 0.18,\\
    \lambda &= \frac{\gamma}{0.8}.
\end{align*}
The parameter $\gamma$ is the aspect ratio at a particular scale, $\theta$ is the orientation $\in$ [$0\,^{\circ}$, $45\,^{\circ}$, $90\,^{\circ}$, $135\,^{\circ}$], $\sigma$ represents the effective width (=0.3 in our case), $\lambda$ is the wavelength at a particular scale, and $\rho$ represents the scale.\\
Secondly, all the S1 image responses are computed by applying a two dimensional convolution between the initialized Gabor filters and the input images in the spatial domain. The S1 image responses are so-called: the Gabor features.\\
In fact, all the filters are arranged in 8 bands. There are two filter scales with four orientations at each band.
\\
The S1 layer has a computational complexity of $\mathrm{O}(N^2 M^2)$ where $M \times M$ is the size of the filter and $N \times N$ is the size of the image

{\bf C1 layer:} The C1 units are considered to have larger receptive field sizes and a certain degree of position and scale invariance. For each band, each C1 unit response (image response) is computed by taking the maximum pooling between the gabor features of the two scales at the same orientation. The main role of the maximum pooling function is to subsample the number of the S1 image responses and increase tolerence to stimulus translation and scaling. Then, the pooling over local neighborhood using a grid of size $n\times n$  is performed. From band 1 to 8, the value of $n$ starts from 8 to 22 in steps of two pixels, respectively. Furthermore, a subsampling operation can also be performed by overlapping between the receptive fields of the C1 units by a certain amount $\Delta_s$ (= $\mathrm{4_{band1}, 5_{band2}, \cdots, 11_{band8}}$), given by the value of the parameter C1Overlap. The value C1Overlap = 2 is mostly used, meaning that half the S1 units feeding into a C1 unit were also used as input for the adjacent C1 unit in each direction. Higher values of C1Overlap indicate a greater degree of overlap. This layer has a computational complexity of $\mathrm{O}(N^2 M)$.

{\bf S2 layer:} The original version of HMAX was the \emph{standard model} in which the connectivity from C1 to S2 was considered \emph{hard-coded} to generate several combinations of C1 inputs. The model was not able to capture discriminating features to distinguish facial images from natural images. To improve that, an extended version was proposed (Serre et al., 2005b), and is called \emph{HMAX with feature learning}. In this model, each S2 unit acts as a Radial Basis Function (RBF) unit, which serves to compute a function of the distance between the input and each of the stored prototypes learned during the feature learning stage. That is, for an image patch X from the previous C1 layer at a particular scale, the S2 response (image response) is given by:
\begin{equation}
\mathrm{S2_{out}}= \exp^{\left(-\beta \| X - P_i\|^2\right)},
\end{equation}
where $\beta$ represents the sharpness of the tuning, ${P_i}$ is the $i\text{th}$ prototype and $\|\cdot\|$ represents the Euclidean distance. This layer has a computational complexity of $\mathrm{O}\left(PN^2 M^2\right)$, where $P$ is the number of prototypes.

{\bf C2 layer:} It is considered the layer at which the final invariance stage is provided by taking the maximum response of the corresponding S2 units over all scales and orientations. The C2 units provide input to the VTUs. This layer has a computational complexity of $\mathrm{O}(N^2 MP)$.

{\bf VTU layer:} At runtime, each image in the database is propagated through the four layers described above. The C1 and C2 features are extracted and further passed to a simple linear classifier. Typically, support vector machine (SVM) and nearest neighbor (NN) classifiers are employed.

{\em The learning stage:} The learning process aims to randomly select {\em P} prototypes used for the S2 units. They are selected from a random image at the C1 layer by extracting a patch of size $4 \times 4$, $8 \times 8$, $12 \times 12$, or $16 \times 16$ at random scale and position (Bands 1 to 8). For an $8 \times 8$ patch size for example, it contains 8 $\times$ 8 $\times$ 8 = 512 C1 unit values instead of 64. This is expected since for each position, there are units representing each of the four orientations [$0\,^{\circ}$, $45\,^{\circ}$, $90\,^{\circ}$, $135\,^{\circ}$].

\section{\uppercase{S1 Layer Approximations}}\label{s:S1}
\noindent At the S1 layer, several approximations are investigated in order to increase the efficiency of the original HMAX model in terms of accuracy and computational complexity. Each approximation has been evaluated independently using SVM and NN classifiers.

\begin{figure*}[t]
  \centering
  \includegraphics[scale=0.7]{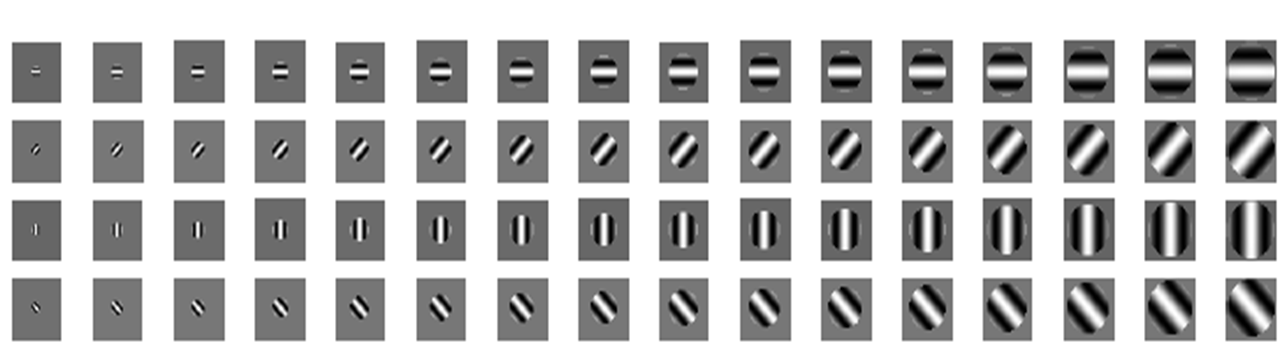}\\
  \caption{64 Gabor filters (16 scales in steps of two pixels [7 $\times$ 7 to 37 $\times$ 37] $\times$ 4 orientations [$0\,^{\circ}$, $45\,^{\circ}$, $90\,^{\circ}$, $135\,^{\circ}$])}\label{f:64_Gabor_filters}
\end{figure*}

\subsection{Combined Image-based HMAX using 2-D Gabor filters}\label{s:2D_Gabor}
\noindent In this approximation, all unimportant information such as illumination and expression variations are eliminated from the image and hence its salient features become richer \citep{x13}. To achieve this, four main steps are applied to the original image A of size $ h\times a$:

\noindent{\textbf{Step 1} -- \emph{Adaptive Histogram Equalization}:} In order to handle the large intensity values to some extent, adaptive histogram equalization is applied to the original image A:
 \begin{equation}
\mathrm{Adapted\_Image} = \mathrm{AdaptHistEq(A)}
\end{equation}
\\
\noindent{\textbf{Step 2} -- \emph{SVD Decomposition}:} Singular value decomposition (\emph{SVD}) is applied to the image after equalization. The concept behind  \emph{SVD} is to break down the image into the product of three different martices as:
\begin{equation}
\emph{SVD}(\mathrm{Adapted\_Image}) = \mathbf{L} \times  \mathbf{D} \times \mathbf{R^T}
\end{equation}
where $\mathbf{L}$ is the orthogonal matrix of size $ h\times h$, $\mathbf{R^{T}}$ is the transpose of an orthogonal matrix $\mathbf{R}$ of size $a\times a$ and $\mathbf{D}$ is the diagonal matrix of size $h \times a$.\\
This decomposition helps the computations to be more immune to numerical errors, as well as to expose the substructure of the original image more clearly and orders their elements from most amount of variation to the least. \\ \\
\noindent{\textbf{Step 3} -- \emph{Reconstruction Image}:} According to the values of $\mathbf{L}$, $\mathbf{D}$ and $\mathbf{R}$, the reconstructed image is computed as follows:
 \begin{equation}
\mathrm{Reconstructed\_Image} = \mathbf{L} * \mathbf{D}^\alpha \mathbf{* R^T},
\end{equation}
where $\alpha$ is a magnification factor that varies between 1 and 2. The idea to have the value of $\alpha$ vary between one and two in order to magnify the singular values of $\mathbf{D}$ is to make them invariant to illumination changes. When $\alpha$ equals to 1, the reconstructed image is equivalent to the equalized image. When $\alpha$ is chosen between ]1~2], then the singular values greater than unity will be magnified. Thus, the combination between the reconstructed image and the equalized image will be a fruitful step to making the model more robust against illumination and expression variations.

\noindent Interestingly, when the singular values are scaled in the exponent, a non-linearity is introduced. Therefore for a specific database (Caltech101 for example), scaling down the magnification factor $\alpha$ may be helpful.\\ \\
\noindent{\textbf{Step 4} -- \emph{Combined Image}:} The combined image is produced by simply combining the reconstructed image and the equalized image as shown in Figure 2, using a combination parameter $c$ which varies between 0 and 1.\small
\begin{equation}
\mathrm{I_{Comb}} = \frac{\mathrm{Adapted\_Image} + (c * \mathrm{Reconstructed\_Image})} {1 + c}
\end{equation}\normalsize
By applying this approximation, the computations in this layer become faster as shown in Figure 5 since only the significant information are used for recognition. In addition, the approximation can significantly improve the model's accuracy. It can be explained by the fact that when the model uses a challenge database such as Calech101 or Caltech256 in which there are a lot of unimportant information such as illumination and expression variations, it will be interesting to exploit only the most important features in the images in order to make the recognition easier and more robust where the accuracy is increased by 10\% using SVM while by more than 13\% when using NN classifier. There are no related works yet that approximate the S1 layer.

\begin{figure*}[t]
  \centering
  \includegraphics[scale=0.62]{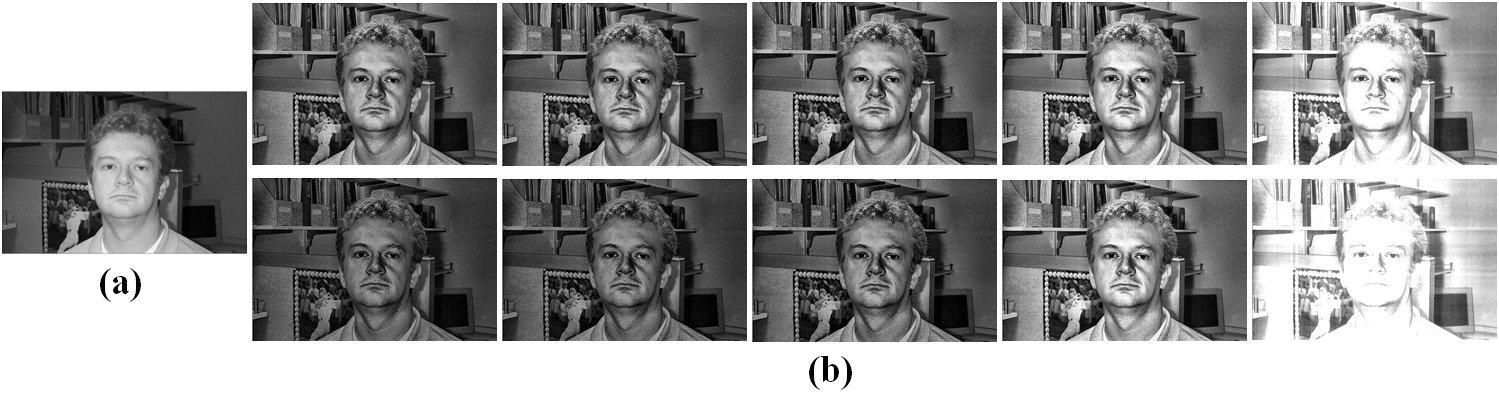}\\
  \caption{(a) The original image and (b) Combined images using $\alpha$ = 0.25, 0.5, 0.75, 1 and 1.25, respectively. $c$ is equal to 0.25 and 0.75 on the top and bottom, respectively}\label{f:combined_images}
\end{figure*}

\subsection{Combined Image-Based HMAX using Separable Gabor Filters}\label{s:seperable_Gabor}
\noindent In this approximation, all the combined images of the previous approximation are convolved with 64 Gabor filters in a separable manner ($G(x,y)=f(x)g(y)$), instead of just performing the 2-D convolution. In this case, the Gabor features are computed using two 1-D convolutions corresponding to convolution by $f(x)$ in the x-direction and $g(y)$ in the y-direction. Based on the definition of separable 2-D filters, the Gabor filters are parallel to the image axes ($\theta=k\pi/2$). In order to be applied to an image along diagonal directions, they have been extended to further work with $\theta=k\pi/4$. The main issue of these techniques is that they will not work with any other desired direction. To handle this problem, eq. (1) can be rewritten using the isotropic version ($\gamma$ = 1, circular) in the complex domain \citep{x9}. In this case,  $ u^2 + v^2 = (xcos\theta + ysin\theta )^2 + (-xsin\theta + ycos\theta)^2$ = $x^2 + y^2.$
\begin{align*}
\mathrm{G}(x,y) &\!=\! e^{-\frac{x^2 + y^2}{2\sigma^2}}\!\times\!\cos\left(\frac{2 \pi}{\lambda} (x\cos(\theta) \!+\! y\sin(\theta)\right)\\
    &= \mathrm{Re}{(f(x)g(y))}
\end{align*}
where
\begin{align*}
    f(x) &= e^{-\frac{x^2}{2\sigma^2}} \times e^{ix\cos(\theta)},\\
    g(y) &= e^{-\frac{y^2}{2\sigma^2}} \times e^{iy\sin(\theta)}.
\end{align*}

Finally, the convolution using this approximation can therefore be expressed as:
\begin{equation}
\mathrm{I_{Comb}} * \mathrm{G}(x,y) =\mathrm{I_{Comb}}(x,y) * f(x) * g(y)
\end{equation}
By exploiting the separability of Gabor filters and convolving them with the original image, the computational complexity is reduced from $\mathrm{O}(N^2 M^2)$ to $\mathrm{O}(tN^2 M)$ where $t$=8 due to complex valued arithmetic. But since in this approximation, the separable Gabor filters are convolved with the combined image  $\mathrm{I_{Comb}}$, the complexity is being more reduced since only the significant information are used for recognition. The accuracy is not increased by more than 10.5\% for SVM (between 10.4\% and 10.5\%) while is increased by more than 14\% for the NN classifier.

\section{\uppercase{C1 Layer Approximations}}\label{s:C1}
\noindent Concerning the C1 layer, a pooling between the S1 responses over scales within each band is performed by simply taking the maximum response between them. By testing what can be the result of the minimum pooling that has not been exploited at this layer, it was noticed that all the minimum scales values are very close to their corresponding maximum ones. Some of them are equal, otherwise the most of minimum scales values are not smaller more than 6 or 7\%. As such, it will be important to further consider some of the minimum scales values when taking the maximum pooling. In other words, some of the minimum scales values can be exploited in addition to the maximum ones in order to increase the model's accuracy. But the remaining question to be solved is "How to take advantage of minimum and maximum scales values at the same time". So that under a specific conditions, some of the minimum scales values can be embedded into their corresponding maximum ones. The easiest way to achieve that is to apply the embedding in the additive domain. A general scheme of this approximation is shown in Figure 3. In this figure, two S1 image responses $\mathrm{I_{1}}$ and $\mathrm{I_{2}}$ of the same orientation at the first band (band1) are considered and which are belong to the filter scale 7 and 9, respectively. The circles shown within the images correspond to their pixels. In step 1, the maximum pooling (max function) is performed between $\mathrm{I_{1}}$ and $\mathrm{I_{2}}$. The pixels of the resulting image correspond to the maximum scales values (shown with blue circles). In step 2, the minimum pooling (min function) is performed between $\mathrm{I_{1}}$ and $\mathrm{I_{2}}$. The pixels of the resulting image correspond to the minimum scales values (shown with violet circles) that are then embedded into their corresponding maximum ones in the additive domain under specific conditions as shown in step 3. In other words, each minimum scale value is added into the maximum one that has the same ($x$, $y$) coordinates. $w$ is the weight of the embedding.

\begin{figure*}[t]
  \centering
  \includegraphics[scale=0.59]{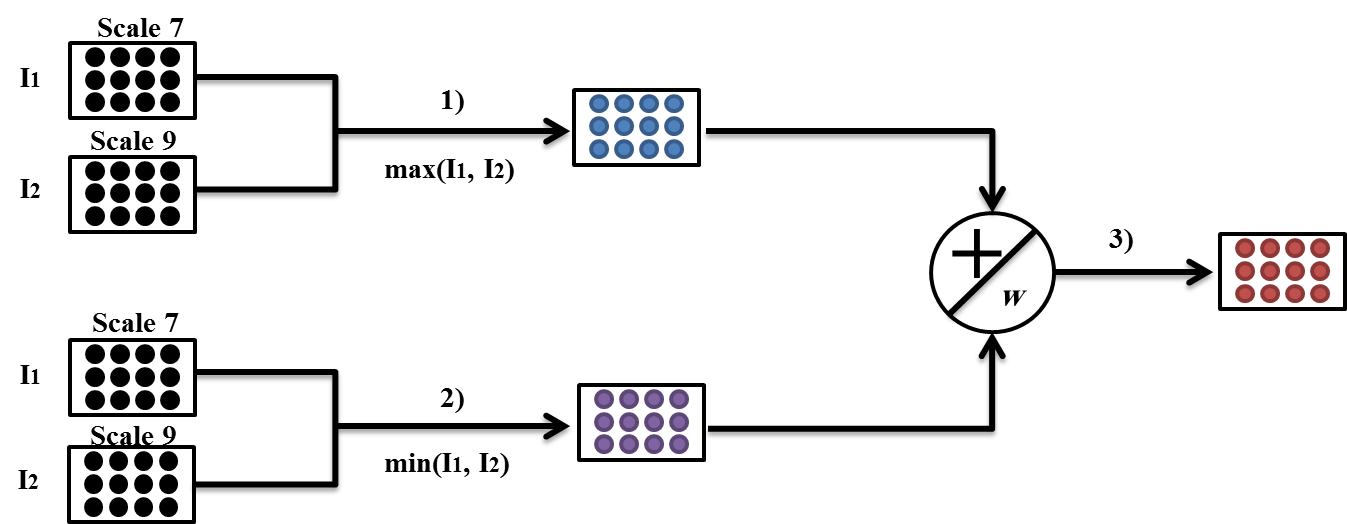}\\
  \caption{Scheme example of the C1 approximation}\label{f:combined_images}
\end{figure*}

\noindent{\textbf{Embedding in the additive domain}:} This kind of embedding is very straightforward to implement since the minimum scales values (after applying the minimum pooling over scales within each band) can be directly embedded into their corresponding maximum values by simply using the addition operator.\\
Generally, the embedding process at a particular pixel coordinate $(x,y)$ in the additive domain can be expressed as:
\begin{equation}
\mathrm{I_{Embed}}(x,y) = \mathrm{max}_{\mathrm{scale}}(x,y) + w * \mathrm{min}_{\mathrm{scale}}(x,y),
\end{equation}
where $\mathrm{I_{Embed}}(x,y)$ represents the final result after the embedding process, $\mathrm{max}_{\mathrm{scale}}$ is the maximum scale value, $\mathrm{min}_{\mathrm{scale}}$ is the minimum scale value, and $w$ $\in$ [0, 1] represents the weight of the embedding.
\\
Two different conditions are considered to embed the minimum scales values into their corresponding maximum ones:\\
{\em Condition 1:} At each band, after computing the maximum pooling over scales of the same orientation, the minimum pooling is also performed and then all the minimum scales values are embedded into their corresponding maximum ones. In this case, $w$ is set to 1.\\
{\em Condition 2:} Each minimum scale value is embedded if and only if its corresponding maximum value belongs to the interval [0\%~5\%[. The values within the interval specifies how much a maximum scale value is greater than its corresponding minimum one. In fact, the interval [0\%~5\%[ is divided into two groups: [0\%~2\%[ and [2\%~5\%[, and two distinct sub-conditions are thus considered:
\begin{itemize}
  \item {\em Sub-condition 1:} The embedding is performed by setting $w$ to 1 for [0\%~2\%[ and 0.5 for [2\%~5\%[.
  \item {\em Sub-condition 2:}~The embedding is performed by setting $w$ to 0.5 for [0\%~2\%[ and 0.1 for [2\%~5\%[.
\end{itemize}

The accuracy is not increased by more than 1\% in all conditions when SVM is used, while the opposite for NN classifier. However, the computational complexity at this layer is slightly increased due to the embedding process.

\section{\uppercase{S2 Layer Approximations}}\label{s:S2}
\noindent At the S2 layer, the focus is to enhance the manner by which all the prototypes are selected during the feature learning stage. In the original model, $P$ prototypes are randomly selected from the training images at the C1 layer. If more than $P$ prototypes are used, the model's accuracy will increase at the expense of additional computational complexity. That is why our motivation is to learn the same number of prototypes $P$ but in an efficient way in order to decrease the model's false classification rate while keeping the same computational complexity.

In order to achieve this, clustering is exploited, which is considered one of the most important research areas in the field of data mining. It aims to divide the data into groups, (clusters) in such a way that data of the same group are similar and those in other groups are dissimilar. Clustering is considered useful to obtain interesting patterns and structures. That is why, one of the existing clustering algorithms, more specifically the Partitioning Around Medoid (PAM) clustering algorithm \citep{x14} has been exploited in this approximation to generate the prototypes.

Furthermore, one of the important issues to consider, is the redundancy of some prototypes especially those selected from the homogeneous areas of the image (prototypes' pixels are being equal to zero). That is why, our contribution also aims to generate a non-redundant $P$ prototypes and force the model not to generate any unimportant prototype. Accordingly, each of the selected prototypes will be important and aims to increase the model's accuracy.

PAM is characterized by its robustness to the presence of noise and outliers. Its complexity is defined by $\mathrm{O}(i(b-q)^2)$ where $i$ is the number of iterations, $q$ is the number of clusters, and $b$ represents the total number of objects in the data set.

To generate 2000 prototypes in a more efficient way and use them in our model instead of the traditional ones, the PAM algorithm is performed and it consists of 6 different steps:

\noindent{\textbf{Step 1} --5 medoids of 4 $\times$ 4 pixels at four orientations of each training category (total of 30 images) from the total 102 categories are randomly initialized.\\

\noindent{\textbf{Step 2} --For each category, the Frobenius distance between each of the C1 response of each image with all the selected medoids is then computed in order to associate each data image to the closest medoid.\\

\noindent{\textbf{Step 3} --For a random cluster, a non-medoid image patch is randomly selected in order to be swaped with the original medoid of the cluser in which the non-medoid is selected.\\

\noindent{\textbf{Step 4} --steps 2 and 3 are repeated until the total cost of swapping becomes greater than zero. The total cost of swapping can be defined as follows:
\begin{equation*}
\mathrm{Cost_{swapping} = Current~Total~Cost~-~Past~Total~Cost}
\end{equation*}

\noindent{\textbf{Step 5} --All the previous steps are also performed for all the other remaining three sizes of the medoids (8 $\times$ 8, 12 $\times$12 and 16$\times$16) in order to have a total of 20 medoids in each category.\\

\noindent{\textbf{Step 6} --Finally, a total of 2040 medoids are being selected to be used as prototypes. 10 prototypes are dropped from each size in order to end up with only 2000 prototypes.
\\
\\
This algorithm is complex since there are six steps to perform in order to generate the prototypes. But in fact, the run of the HMAX model relies on two parts. The first part is responsible to generate and reserve all the necessary prototypes by only running the first two layers S1 and C1. The second part consists of running the whole model and use the prototypes that have been generated and reserved for the S2 layer.
Interestingly, the complexity of the model depends only on the second part, which means that the large complexity of our algorithm does not affect the computational complexity of the model, more precisely, of the S2 layer. That is why, the computational complexity at the S2 layer of our model remains $\mathrm{O}\left(PN^2 M^2\right)$, where $P$ is the number of prototypes.\\
By applying this approximation, the accuracy of the model incerases by 0.68\% approximately using the SVM classifier.

\section{\uppercase{Experimental Results}}\label{s:results}
\noindent The proposed optimizations at the S1, C1 and S2 layers were implemented using MATLAB in order to evaluate their accuracy and computational complexity using experimental simulations. The S1, C1 and S2 approximations were evaluated using the Caltech101 database, which contains a total of 9,145 images split between 101 distinct object categories in addition to a background category. All the results of our approximations were the average of 3 independent runs. For each run, the following steps were performed:
\begin{enumerate}
  \item A set of 30 images are randomly chosen from each category for training, while all the remaining images are used for testing. All the images are normalized to 140 pixels in height and the width is rescaled accordingly so that the image aspect ratio is preserved.
  \item C1 sub-sampling ranges do not overlap in scales.
  \item The prototypes are learned at random scales and positions. They are extracted from all the eight bands.
  \item C2 vectors are built using the training set.
  \item Training applied using both SVM and Nearest-Neighbor classifiers.
  \item C2 vectors for the test set are built, and then the test images are classified.
\end{enumerate}

\subsection{Performance of SVM and NN}
\noindent  The performance of both SVM and NN are performed on the face category extracted from the caltech101 database and which contains 435 face images. The images were rescaled to 160 $\times$ 160 pixels, the C1 sub-sampling ranges overlap in scale (C1Overlap = 2) and the prototypes are chosen only from Bands 1 and 2. The classifiers were trained with $n=50$, 100, 150, 200, 250, 300, 350 and 400 positive examples and 50 negative examples from the background class, while they are tested with all the remaining positive examples and 50 examples from the negative set as shown in Table 1 and Figure 4. 1000 prototypes (250 patches) $\times$ (4 sizes) are used in the S2 layer.

The results show that the accuracy decreases when the number of training becomes greater than 300. This is expected because the data becomes unbalanced.
\begin{table}[hbtp]
\centering
\caption{Simulation results for SVM and NN on face category}
\begin{tabular}{ | l | l | l |}
\hline
\bf Positive training & \bf ~~SVM & \bf ~~~NN\\
\hline
~~~~~~~~~~~50 &  92.325\% &  52.903\% \\
\hline
~~~~~~~~~~100 &  95.678\% &  79.578\%\\
\hline
~~~~~~~~~~150& 96.656\% & 88.240\%\\
\hline
~~~~~~~~~~200  & 97.018\% & 90.658\%\\
\hline
~~~~~~~~~~250 & 97.075\% & 92.181\%\\
\hline
~~~~~~~~~~300 & 97.165\% & 92.634\%\\
\hline
~~~~~~~~~~350 & 96.019\% & 91.480\%\\
\hline
~~~~~~~~~~400 & 94.558\% & 87.649\%\\
\hline
\end{tabular}
\label{t:svm_vs_NN}
\end{table}
$\\$
\begin{tikzpicture}
        \begin{axis}[%
            axis x line=bottom,
            axis y line=left,
            xlabel=$Number~of~positive~training$,
            ylabel=$Correct~classification~rate~(\%)$,
            width =6.9cm,
            legend pos= south east]
            \addplot[mark=none,solid,thick] coordinates {(50, 92.325) (100, 95.6775) (150, 96.656) (200, 97.0175) (250, 97.075) (300, 97.165) (350, 96.01875) (400, 94.5575)};
         \addplot[mark=none,dashed,thick] coordinates {(50, 52.9025) (100, 79.5775) (150, 88.24) (200, 90.6575) (250, 92.18125) (300, 92.63375) (350, 91.48) (400, 87.64875)};

\legend{SVM, NN};
        \end{axis}

    \end{tikzpicture}
\begin{center}
{\footnotesize Figure 4: SVM and NN accuracies on face category}
\end{center}

\subsection{Evaluations at the S1 layer}
At this layer, the computational complexity and correct classification rates (accuracies) for each of the proposed approximations ($\bf Approx$) are compared to the baseline model.
\begin{itemize}
  \item {\bf Approx1:}~Combined~Image-based~HMAX using 2-D Gabor filters.
  \item {\bf Approx2:}~Combined~Image-based~HMAX using separable Gabor filters.
\end{itemize}

\noindent In order to compute the speed of the approximations at this layer, the total time complexity of the S1 layer is measured on a specific face image from the face category. All the evaluations were done on a core i7 2.4 GHZ machine. The simulations were repeated five times. Figure 5 illustrates an average of the results. It shows that both Approx1 and Approx2 are faster than the baseline (blue curve) for all the tested image sizes. It has been noticed that for an image of size between 100x100 and 160x160, Approx1 is always faster than Approx2. For example, Approx1 is faster than Approx2 by 3.23\% for an image of size 100x100. For other image sizes greater than or equal to 160x160, Approx2 always shows lower timing than Approx1. For example, for an image of size 160x160, Approx1 is faster than the baseline by 2.95\% while by 3.42\% for Approx2. For an image of size 256x256, Approx1 is faster than the baseline by 6.29\% while by 12.56\% for Approx2.

$\\$
\begin{tikzpicture}
\begin{axis}[
  symbolic x coords={100x100,160x160, 256x256, 512x512, 640x480},
  xtick={100x100,160x160, 256x256, 512x512, 640x480},
  xticklabel style={rotate=-90},
  ylabel={Time (s)},
width = 6.9cm,
 legend style={at={(0.05, 0.6)},anchor=south west}];
\addplot coordinates {(100x100, 0.1698) (160x160, 0.2422) (256x256, 0.4268) (512x512, 1.4764) (640x480, 1.5844)};

\addplot coordinates {(100x100, 0.1360) (160x160, 0.2127) (256x256, 0.3639) (512x512, 1.1378) (640x480, 1.2350)};

\addplot coordinates {(100x100, 0.1683) (160x160, 0.2080) (256x256, 0.3012) (512x512, 1.0087) (640x480, 1.1564)};

\legend{Baseline, Approx1, Approx2};

\end{axis}
\end{tikzpicture}
\begin{center}
{\footnotesize Figure 5: Timing comparison (in s)}
\end{center}

In order to assess the correct classification rates, both SVM and NN classifiers were used. The average accuracies of Approx1 under different values of $\alpha$ and $c$ are shown in Table 2. From all the following experiments, 2000 prototypes (500patches) $\times$ (4 sizes) are used and all the images were rescaled to 140 in height. Recall that C1 sub-sampling ranges do not overlap in scales and the prototypes are extracted from all the eight bands. The performance of the original model reaches
39\% and 21.2\% when using 30 training examples per class averaged over 3 repetitions under SVM and NN, respectively. Table 2 proves our significant contribution at the S1 layer especially for $\alpha$ = 0.75 and $c$ = 0.25 where the accuracy is increased by 10.02\% and 13.811\% using SVM and NN, respectively.

\begin{table}[hbtp]
\centering
\caption{Classification accuracies of Approx1 approximation}\label{t:approx1_classif}
\begin{tabular}{ | l | l | l | l |}
\hline
Approx1 & Classifier  & $c$ = 0.25 & $c$ = 0.75\\
\hline
$\alpha$ = 0.25 & ~~~SVM & 34.3652\% & 33.5927\% \\
\cline{2-4}
 & ~~~~NN & 20.284\% & 20.283 \\
\hline
$\alpha$ = 0.5 & ~~~SVM & 45.201\% & 45.158\%  \\
\cline{2-4}
& ~~~~NN & 31.23\% & 31.231 \\
\hline
\bf $\alpha$ = 0.75 & \bf ~~~SVM & \bf 49.02\% & 48.269\%  \\
\cline{2-4}
& \bf ~~~~NN & \bf 35.011\% & 34.455\% \\
\hline
$\alpha$ = 1 & ~~~SVM & 47.739\% & 47.739\%  \\
\cline{2-4}
& ~~~~NN & 31.362\% & 31.362\% \\
\hline
$\alpha$ = 1.25 & ~~~SVM & 39.3531\% & 40.743  \\
\cline{2-4}
& ~~~~NN & 23.998\% & 24.08\% \\
\hline
\end{tabular}
\end{table}

Figure 6 illustrates the results shown in Table 2. It shows 4 different curves. The red and blue solid curves represent the accuracy values for $c$ = 0.25 under SVM and NN, respectively. While the red and blue dashed curves are for $c$ = 0.75 under SVM and NN, respectively.

Finally, the separability of Gabor filters is exploited and applied to the combined image with $\alpha = 0.75$ and $c = 0.25$. Approx2 shows an accuracy equal to $49.471\%$ and $35.372\%$ for SVM and NN, respectively.
$\\$
$\\$
\begin{tikzpicture}
\begin{axis}[
  symbolic x coords={0.25, 0.5, 0.75, 1, 1.25},
  xtick={0.25, 0.5, 0.75, 1, 1.25},
  xticklabel style={rotate=0},
 xlabel=$\alpha$,
  ylabel={accuracy (\%)},
  width = 6.9cm,
  legend pos= south east]
  \addplot [mark=o,solid,red, thick] coordinates {(0.25, 34.3652) (0.5, 45.201) (0.75, 49.02) (1, 47.739) (1.25, 39.3531)};
\addplot [mark=o,solid,blue, thick] coordinates {(0.25, 20.284) (0.5, 31.23) (0.75, 35.011) (1, 31.3618) (1.25, 23.998)};

  \addplot [mark=o,dashed,red, thick] coordinates {(0.25, 33.5927) (0.5, 45.158) (0.75, 48.269) (1, 47.739) (1.25, 40.743)};
\addplot [mark=o,dashed,blue, thick] coordinates {(0.25, 20.283) (0.5, 31.231) (0.75, 34.455) (1, 31.362) (1.25, 24.08)};
\end{axis}
\end{tikzpicture}
\\
{\footnotesize Figure 6:  Approx1 accuracies under different values of $\alpha$ and $c$}

\subsection{Evaluations at the C1 layer}
Figure 7 shows the average accuracies of the SVM (blue x points) and NN (red x points) on several C1 optimization options ($\bf Opt$). The accuracy of the model is increased a little bit when three cases of the additive method are applied. For example, using SVM, the accuracy is increased by 0.577\%, 0.607\%, 0.843\% on Opt1, Opt2 and Opt3, respectively. On the other hand, the increase is 0.846\%, 1.88\%, 1.85\% using NN.
\begin{itemize}
    \item  {\bf Opt1:}~Embedding all pixels ($\alpha=1$).
    \item  {\bf Opt2:}~$[0\%, 2\%[$, ($\alpha =0.5$); $[2\%, 5\%[$, ($\alpha=0.1$)
    \item  {\bf Opt3:}~$[0\%, 2\%[$, ($\alpha =1$); $[2\%, 5\%[$, ($\alpha=0.5$)
\end{itemize}
$\\$
\begin{figure}
  \centering
\begin{tikzpicture}
\begin{axis}[
  symbolic x coords={Baseline, Opt1, Opt2, Opt3},
  xtick={Baseline, Opt1, Opt2, Opt3},
  xticklabel style={rotate=-90},
width = 6.9cm,
only marks,
 legend style={at={(0.74, 0.65)},anchor=south west},
  ylabel={Accuracy (\%)}]
\addplot[mark=x,blue, thick] coordinates {(Baseline, 39) (Opt1, 39.577) (Opt2, 39.607) (Opt3, 39.843)};
\addplot[mark=x,red,thick]  coordinates{(Baseline, 22.2) (Opt1, 22.046) (Opt2, 23.08) (Opt3, 23.05)};
\legend{SVM, NN};
\end{axis}
\end{tikzpicture}
{\footnotesize Figure 7:  Average accuracies of C1 approximations}
\end{figure}

\subsection{Evaluations at the S2 layer}
Table 3 shows the average accuracy of the SVM classifier based on the S2 approximation.
\begin{table}[hbtp]
\centering
\caption{Accuracy of S2 approximation}
\label{t:S2_aprrox}
\begin{tabular}{ | l | l | l |}
\hline
\bf Approximation & ~~~~~~~~~~\bf SVM\\
\hline
~~~~~Baseline & ~~~~~~~~~~39\% \\
\hline
~~~~~~~PAM & 39.68\% (+0.68\%) \\
\hline
\end{tabular}
\end{table}

This approximation has a big advantage on the model since the selected prototypes are non-redundant and generated in more intelligent way. Therefore, each prototype serves to slightly increase the accuracy. The accuracy of the model is increased approximately by 0.68\%.
\subsection{Combined Classification Accuracies}
Figure 8 shows the average accuracies of SVM on the combination of the approximations "Approx2" + "Opt3" + "PAM". Our model shows an accuracy equal to 51\% when using only 2000 prototypes while it shows 53.8\% when using higher number of protoypes (4080) as used by \citet{x6}, \citet{x8}, \citet{x10}, \citet{x11}.

\begin{figure}[hbtp]
  \centering
  \begin{tikzpicture}
\begin{axis}[
ybar,
  symbolic x coords={Serre et al. (2005b), Holub and Welling (2005), Our model, Our model', Mutch and Lowe (2006), Grauman and Darrell (2005)},
  xtick={Serre et al. (2005b), Holub and Welling (2005), Our model, Our model', Mutch and Lowe (2006),  Grauman and Darrell (2005)},
  xticklabel style={rotate=-90},
nodes near coords,
  ylabel={Accuracy (\%)},
width = 6.9cm,
  legend pos= north east]

\addplot [black] coordinates{(Serre et al. (2005b), 42) (Holub and Welling (2005), 43) (Our model, 51) (Our model', 53.8) (Mutch and Lowe (2006), 56) (Grauman and Darrell (2005), 58.2)};



\end{axis}
\end{tikzpicture}
\begin{center}
{\footnotesize Figure 8:  The accuracies of the final models}
\end{center}
\end{figure}

\section{Discussion and Future Work}
In this study, the complexity of all the five different layers of the original model of object recognition in the visual cortex, HMAX, is presented. Different approximations were added to the first three layers S1, C1 and S2.\\
The results showed that removing all unimportant information such as illumination, expression variations and occlusions, to be a fruitful approach to improving performance. The idea behind separability of Gabor filters has been also exploited in order to be applied on the combined images generated after keeping only the important features for recognition. The change of the main concept at the C1 layer is further applied by exploiting the advantage of some of the minimum scales values and using them to be embedded into the extracted maximum scales values. The accuracy was slightly increased when the embedding process has been applied using the additive method.  Our model serves also to always use an intelligent version of selected prototypes at the S2 layer in order to remove all the possibilities of having an unimportant prototype aiming to decrease the model's accuracy.

As for future enhancements, a likely first step would be to attempt to extend our work to the HMAX model in color mode. In addition, several new approximations will be applied and tested on more challenging databases.


{\small
\bibliography{test}}

\vfill
\end{document}